\documentclass[a4paper,fleqn]{cas-dc}

\usepackage[authoryear,longnamesfirst]{natbib}
\usepackage{subcaption}
\def\tsc#1{\csdef{#1}{\textsc{\lowercase{#1}}\xspace}}
\tsc{WGM}
\tsc{QE}
\begin{document}
\let\WriteBookmarks\relax
\def\floatpagepagefraction{1}
\def\textpagefraction{.001}

\shorttitle{Deep Learning based Direct Leaf Area Estimation}    

\shortauthors{Namal, Oula, Wen, Yi}  

\title [mode = title]{Deep Learning-Based Direct Leaf Area Estimation using Two RGBD Datasets for Model Development.}  

\author[1]{Namal Jayasuriya}%[<options>]

\cormark[1]


\ead{N.Jayasuriya@westernsydney.edu.au}

% \ead[url]{<URL>}

\credit{Conducted research design, experiments, data collection, data preperation, programming, and prepared the manuscript}

\affiliation[1]{organization={Western Sydney University},
            addressline={Hawkesbury Institute for the Environment}, 
            city={Richmond},
%          citysep={}, % Uncomment if no comma needed between city and postcode
            postcode={2753}, 
            state={NSW},
            country={Australia}}

% ------------ Author 2 ----------
\author[2]{Yi Guo}%[<options>]

% Footnote of the second author

% Email id of the second author
\ead{Y.Guo@westernsydney.edu.au}


\credit{Equally contributed in research design and manuscript preparation}

\affiliation[2]{organization={Western Sydney University},
            addressline={Centre for Research in Mathematics and Data Science}, 
            city={Parramatta},
%          citysep={}, % Uncomment if no comma needed between city and postcode
            postcode={2150}, 
            state={NSW},
            country={Australia}}

% ------------ Author 3 ----------
\author[3]{Wen Hu}%[<options>]

% Footnote of the second author
% \fnmark[3]

% Email id of the second author
\ead{wen.hu@unsw.edu.au}


\credit{Equally contributed in research design and manuscript preparation}

\affiliation[3]{organization={The University of New South Wales},
            addressline={School of Computer Science and Engineering}, 
            city={Sydney},
%          citysep={}, % Uncomment if no comma needed between city and postcode
            postcode={2052}, 
            state={NSW},
            country={Australia}}
            
% ------------ Author 4 ----------
\author[1]{Oula Ghannoum}%[<options>]

% Footnote of the second author
% \fnmark[4]

% Email id of the second author
\ead{O.Ghannoum@westernsydney.edu.au}


\credit{Equally contributed in research design and manuscript preparation}

% -------------------------------------------------------------------------------

\cortext[1]{Corresponding author}


\begin{abstract}
   Estimation of a single leaf area can be a measure of crop growth and a phenotypic trait to breed new varieties. It has also been used to measure leaf area index and total leaf area.
   Some studies have used hand-held cameras, image processing 3D reconstruction and unsupervised learning-based methods to estimate the leaf area in plant images. Deep learning works well for object detection and segmentation tasks; however, direct area estimation of objects has not been explored. This work investigates deep learning-based leaf area estimation, for RGBD images taken using a mobile camera setup in real-world scenarios. A dataset for attached leaves captured with a top angle view and a dataset for detached single leaves were collected for model development and testing.  First, image processing-based area estimation was tested on manually segmented leaves. Then a Mask R-CNN-based model was investigated, and modified to accept RGBD images and to estimate the leaf area. The detached-leaf data set was then mixed with the attached-leaf plant data set to estimate the single leaf area for plant images, and another network design with two backbones was proposed: one for segmentation and the other for area estimation. Instead of trying all possibilities or random values, an agile approach was used in hyperparameter tuning. The final model was cross-validated with 5-folds and tested with two unseen datasets: detached and attached leaves. The F1 score with 90\% IoA for segmentation result on unseen detached-leaf data was 1.0, while R$^2$ of area estimation was 0.81. For unseen plant data segmentation, the F1 score with 90\% IoA was 0.59, while the R$^2$ score was 0.57. The research suggests using attached leaves with ground truth area to improve the results.
\end{abstract}


\begin{highlights}
\item Deep leaning models for segmenting fully visible single leaves and estimating their area without resource- and time-consuming 3D reconstruction.
\item Agile approach of hyperparameter tuning towards a better model design and configurations instead of trying all possible combinations.
\item A case study of mixing two datasets (detached-leaf images with ground truth areas and plant images without ground truths) for model development to investigate in-situ performance.
\item Comparison of traditional 3D reconstruction-based area estimation on segmented leaves and deep learning-based direct area estimation over manually measured leaf area.
\end{highlights}

% Keywords

\maketitle

% % Main text
% \section{}\label{}

% % Numbered list
% % Use the style of numbering in square brackets.
% % If nothing is used, default style will be taken.
% %\begin{enumerate}[a)]
% %\item 
% %\item 
% %\item 
% %\end{enumerate}  

% % Unnumbered list
% %\begin{itemize}
% %\item 
% %\item 
% %\item 
% %\end{itemize}  

% % Description list
% %\begin{description}
% %\item[]
% %\item[] 
% %\item[] 
% %\end{description}  

% % Figure
% \begin{figure}[<options>]
% 	\centering
% 		\includegraphics[<options>]{}
% 	  \caption{}\label{fig1}
% \end{figure}

% \begin{table}[<options>]
% \begin{tabular*}{\tblwidth}{@{}LL@{}}
% \toprule
% \midrule
%  & \\
%  & \\
%  & \\
%  & \\
% \bottomrule
% \end{tabular*}
% \end{table}

% % Uncomment and use as the case may be
% %\begin{theorem} 
% %\end{theorem}

% % Uncomment and use as the case may be
% %\begin{lemma} 
% %\end{lemma}

% %% The Appendices part is started with the command \appendix;
% %% appendix sections are then done as normal sections
% %% \appendix

% \section{}\label{}

% Main text
% ------------------------------------------------------------------------------
\section{Introduction}\label{sec:intro}

Agricultural productivity ultimately hinges on the efficient conversion of light energy into plant biomass, a process primarily driven by photosynthesis in the leaves. Leaf area plays a critical role in this process, directly influencing the plant’s ability to absorb light and convert carbon dioxide into organic carbon compounds. Therefore, accurate measuring of the leaf area is fundamental for assessing plant growth, health, and productivity. It is widely used in breeding programmes to compare varieties~\citep{damatta_exploring_2004,niu_monitoring_2018, phoncharoen_determination_2022}, monitor growth rates~\citep{weraduwage_relationship_2015}, and detect stress factors such as drought~\citep{widuri_relative_2017} or nutrient deficiencies~\citep{hu_reduction_2020}.
Although total leaf area (TLA) and leaf area index (LAI) are commonly used growth estimation metrics and originally based on the single leaf area, which is labour intensive. Hence, indirect alternative methods have been used, such as
Green Area Index (GAI), Projected Leaf Area (PLA), and 3D reconstruction-based total leaf area index, but often fall short due to complexities such as leaf overlap and canopy heterogeneity~\citep{baret_gai_2010, lai_correcting_2022}. ~\cite{de_bock_review_2023} have reviewed LAI measurement in vertical greening systems and shown the lack of methodologies for LAI measurement. They have highlighted the possibility of using 3D data using scanners such as LIDAR. 
In general, these methods focus primarily on aggregate measures, overlooking the importance of the individual leaf area.
Estimation of single-leaf area offers a more detailed and precise approach for monitoring plant growth and health. It provides valuable insight into individual leaf development, which can be aggregated to estimate the total leaf area or detect anomalies at the leaf level. This metric is crucial for vertical farming systems, where leaves grow in overlapping layers, making traditional methods with top-down view less effective. In addition, single-leaf area estimation supports applications such as specific leaf area (SLA) calculations, varietal comparisons, and photosynthetic efficiency assessments.

Traditional methodologies for measuring single-leaf area rely on destructive techniques using leaf area metres LI-3100C~\citep{li-cor_inc_li-3100c_nodate}, which require leaves to be detached from the plant and measured in a laboratory. Portable leaf area machines such as CI-202 from ~\cite{cid_inc_ci-202_nodate} are also available that require an operator and are also not convenient to use as they may require the clamping of leaves. Some researchers have attempted non-destructive approaches, such as using leaf length and width with regression models~\citep{khan_use_2016, teobaldelli_developing_2019}. However, these methods require manual work and are labour intensive for large-scale measurements. 

% Recent advances in computer vision and machine learning offer a promising alternative. 
Traditional image processing-based methods have already been tried for area estimation tasks. For example, ~\cite{yau_portable_2021} developed a system using Kinect V2 for 3D reconstruction and unsupervised learning for leaf segmentation, achieving an R² range of 0.79 to 0.91. This system is used as a handheld device, and images are obtained with a focus on regions of interest (RoI). Then 3D point cloud clustering on 3D reconstructed is used for density-based object segmentation, and manual selection of segmented leaves is used for 3D reconstruction-based area calculation. Depth imaging-based 3D reconstruction and especially 3D point cloud clustering are relatively time-consuming tasks compared to deep learning-based estimations~\citep{jayasuriya_machine_2024}. Review by ~\cite{tian2022application} that discusses image-based crop monitoring methodologies also does not show strong methods for image-based single leaf area estimation while also verifying the limitations of image processing-based methods. We identified the need of focused ROI, manual selection of leaves, and time- and resource-costly 3D reconstruction as barriers towards automated single leaf area estimation for complex crop facilities.

Leaf instances segmentation is the first and essential step for leaf-level area estimation in multi-object images. Mask R-CNN~\citep{he2018maskrcnn} is a popular instance segmentation architecture, and YoLo~\citep{Jocher_Ultralytics_YOLO_2023} is a faster and accurate object detection model that also provides instance segmentation from 2023.
SAM2~\citep{ravi2024sam2segmentimages} is another latest Vision Transformer-based segmentation and tracking architecture, but it requires prompt input of spatial information for initial segmentation. 
Recent research has shown the capabilities of these architectures with 0.5 IOU and YOLO outperforming while Mask R-CNN shows competitive results ~\citep{hu_detection_2023_yolo_mrcnn, fish_segment2024_yolo_mrcnn}. 
However, direct deep learning-based area estimation extending these models has not been investigated.

In this research, we focus on developing a non-destructive, deep learning-based direct and robust method for single-leaf area estimation for moving camera setups in glasshouses. Our approach extends an instance segmentation model to accurately detect, segment, and estimate area of individual leaves from RGBD images of capsicum plants (Figure~\ref{fig:sla-rcnn-N2}). This enables precise area estimation without the need for leaf detachment or complex and time-consuming 3D reconstructions and 3D processing. We leverage two datasets, one with isolated leaves which is easily available for collection of images at different distances and ground truths, and another with entire plants (i.e. attached leaves) to train segmentation for real-world applicaiton and to validate our model, ensuring robustness in real-world scenarios. This work addresses critical gaps in existing methodologies by providing a scalable, relatively accurate, and easy-to-use tool for single-leaf area estimation in greenhouse settings. The following sections outline our data collection methodology, model architecture, training processes, and experimental results, demonstrating the effectiveness of our approach in real-world conditions.

\section{Methodology}\label{sec:method}
\subsection{Data collection and preparation} \label{subsec:data_collection}

\begin{figure*}[t]
  \centering
   \includegraphics[width=1.0\linewidth]{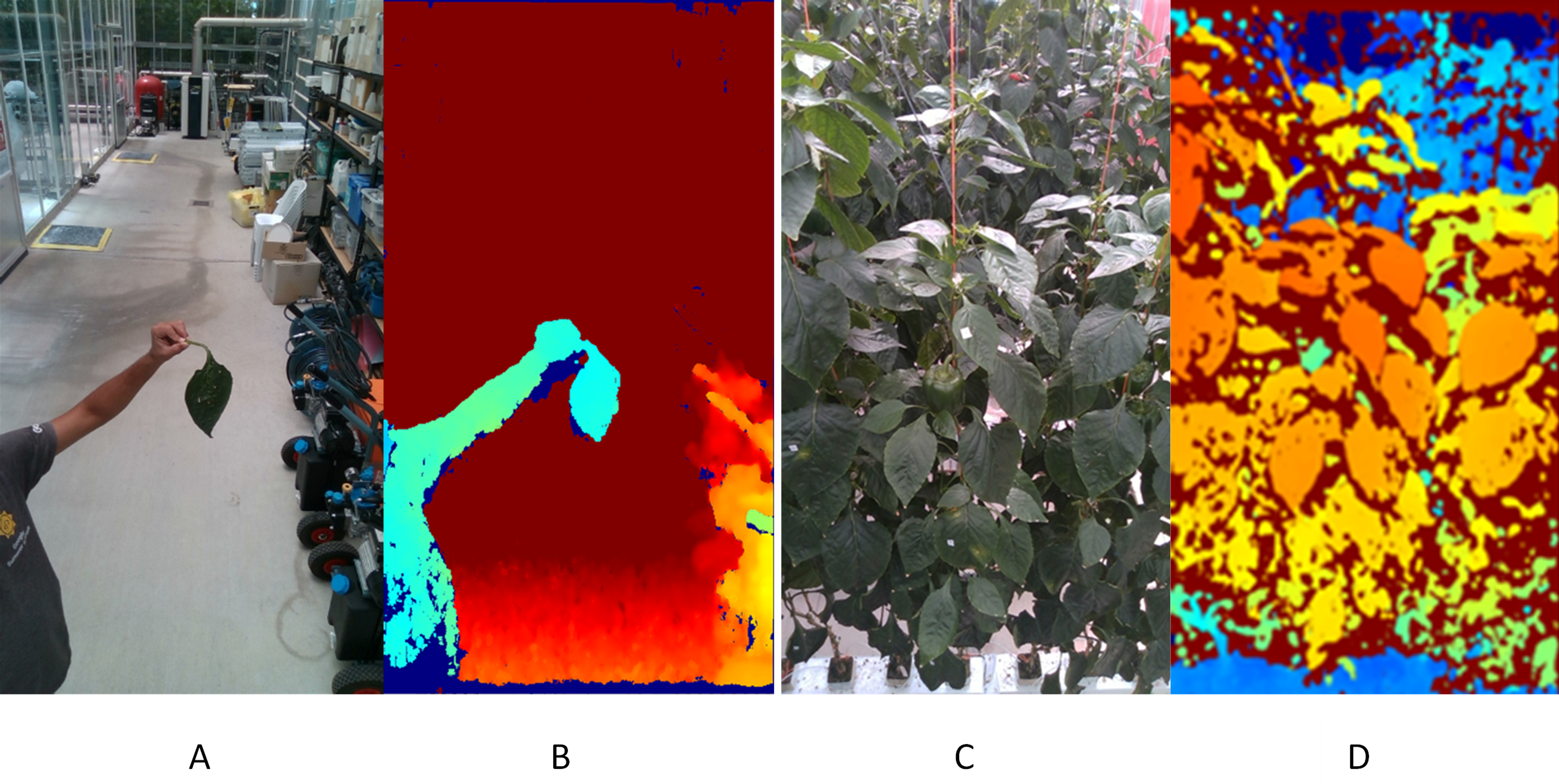}

   \caption{Samples of collected data. A: Images of a detached single-leaf while moving to different distances, B: Corresponding depth image of image A, C: An image for testing leaves on the plant (selected images are labelled with a white colour piece of tape with a label), and D: Corresponding depth image of image C}
   \label{fig:data}
\end{figure*}

This work was carried out using data collected on capsicum crop grown in the glasshouse at National Vegetable Protected Cropping Centre (NVPCC), the Hawkesbury campus of Western Sydney University in Richmond, Australia. 
For training and evaluation, images of detached single leaves were collected 
with ground truths (dataset-1). First, the pruned leaves were selected with diversity (small, medium, and large) in terms of their size. Then, a realsense D415~\citep{intel-realsense-d400-datasheet} RGBD camera was set up at a fixed height in the glasshouse and tilted 30-35 degrees from the vertical axis. RGBD stream was recorded for each individual leaf while it was moving away from the camera (from 0.5 m to 2.5 m) imitating its appearance at different heights, while maintaining similar orientation as it was naturally on the plant (Figure~\ref{fig:data} A and B). Ground truth leaf areas were measured using Li-3100C LI-COR leaf area metre~\citep{li-cor_inc_li-3100c_nodate}. 
For improving segmentation for images of on-plant leaves, plant images from PC4C\_CAPSI\footnote{RGBD data collected weekly at a glasshouse capsicum crop to monitor crop growth.} dataset~\citep{jayasuriya_pc4c_capsi_2024}, collected using Spi-VSTL-V1\footnote{ Mobile camera setup for image data collection at state-of-the-art glasshouse pipe rails system.}~\citep{jayasuriya_spi-vstl_2024} were used (dataset-2). For robust segmentation of leaves at different growth stages, images were selected from 12 time points (2 weeks gap between two) of the growth cycle. For further in-situ testing in the targeted real-world scenario, another small dataset was collected following PC4C\_CAPSI dataset and ground truth areas were collected for the selected leaves (Figure~\ref{fig:data} C and D).

RGBD image data were extracted from recorded streams and pre-processed following the PC4C\_CAPSI dataset. The CVAT~\citep{CVAT_2023} tool was then used to annotate fully visible leaves which were recognised as over 90\% visible to the naked eye. Annotated data were used in coco format, ground truth leaf area was also added and place holder value of -1 was used when ground truth area was not available. The depth scale obtained at image extraction was added to the image information. The data were then split to 5-folds for cross-validation. The size of the training, validation and testing data sets is shown in Table~\ref{tab:dataset_splitting}.

\begin{table}[ht!]
    \centering
    % \small{
    \begin{tabular*}{\linewidth}
    % {lcc}
    {@{\extracolsep{\fill}}p{0.3\linewidth}p{0.3\linewidth}p{0.3\linewidth}@{}}
        \hline
        Splitting    & Leaves on leaf images & Leaves on plant images \\
        \hline
        Training     & 222         & 1059         \\
        Validation   & 58          & 264          \\
        Testing      & 29          & 22           \\
        \hline
    \end{tabular*}%}
    \caption{Data splitting for training, validation, and testing}
    \label{tab:dataset_splitting}
\end{table}

\subsection{Investigation of Image Processing-based Area Estimation }\label{sec:image_processing}

\begin{figure*}[t]
  \centering
   \includegraphics[width=0.6\linewidth]{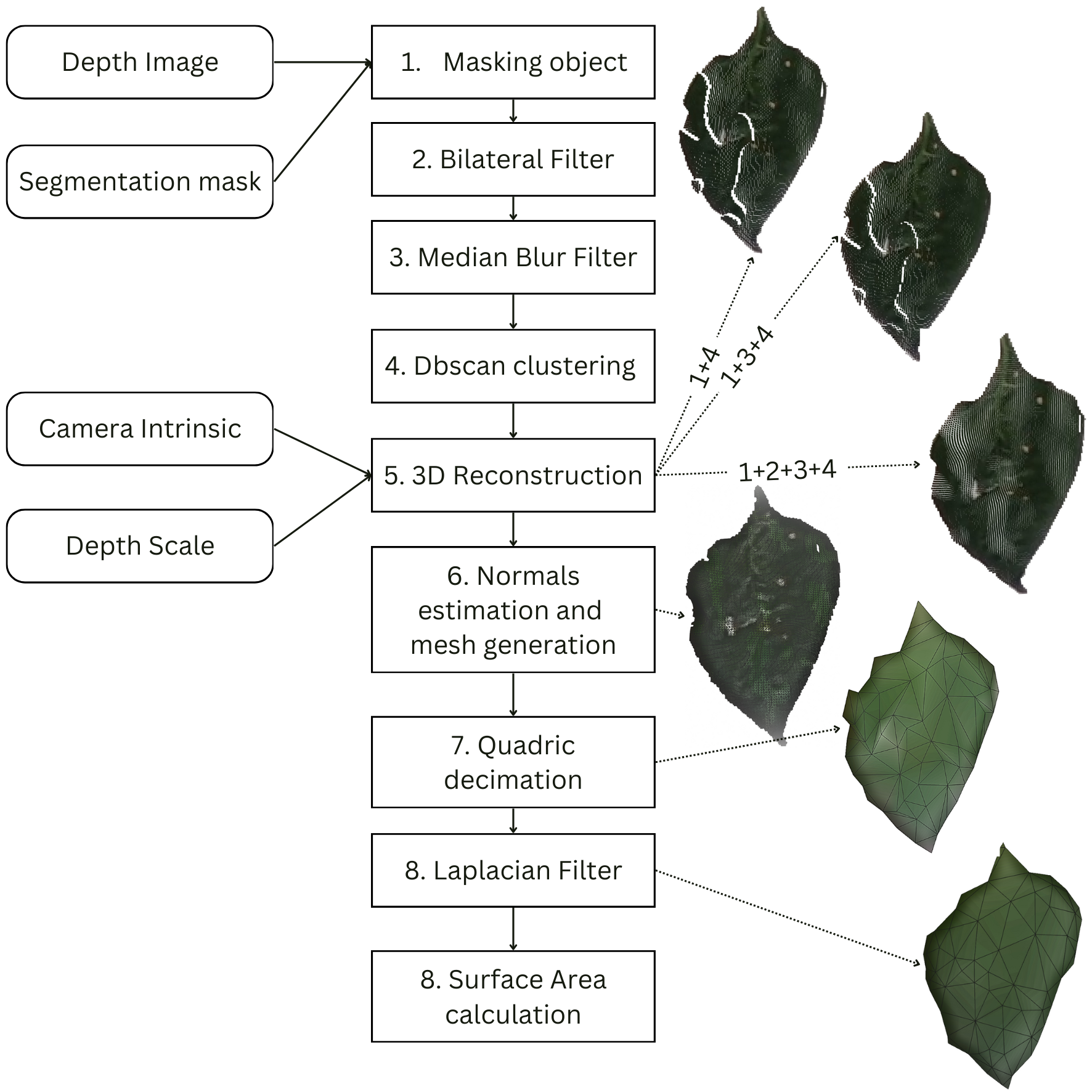}

   \caption{Workflow of image processing-based area calculation of a segmented leaf}
   \label{fig:image-processing based area}
\end{figure*}

Manual annotations of single leaves were used for image processing-based area estimation, serving as a benchmark for deep learning approaches. Figure~\ref{fig:image-processing based area} shows the workflow for this method, illustrating the effects of various filters. Depth images exhibited two types of noise: layered depth maps and impulse noise. Bilateral filter was employed to smooth the layered depth noise. This filter is an edge-preserving filter that uses both spatial gaussian and intensity gaussian respectively for weighting nearby pixels and weighting pixels with similar intensities to calculate weighted averages of nearby pixels. For processing detached-leaf data, the diameter of pixel neighbourhood (d): 10, std of intensity domain (sigmaColor): 150, std of spatial domain (sigmaSpace): 50 were used. The 3D reconstructed leaf with and without bilateral filter in Figure~\ref{fig:image-processing based area} shows its effect on smoothing depth layers. Then the median filter with kernel size of 5 was used to smooth the impulse noise which is also called as salt-and-pepper noise.
%OpenCV (Bradski \& Kaehler, 2000)

After smoothing the depth images, the Open3D library~\citep{Zhou2018open3d} was used to generate the 3D point cloud of a segmented leaf using both RGB and depth images, together with the depth scale and the camera intrinsics. When a segmented leaf was mapped to 3D space, there were some segments appearing outside the leaf due to depth noises for some leaves. The density-based clustering algorithm (DBSCAN) \citep{ester1996dbscan} for 3D point clouds was used to filter out the 3D point cloud of the leaf object. DBSCAN clusters the data points while eliminating outlier points using the neighbourhood radius and minimum number of points for determining a cluster. The largest cluster obtained was selected as the leaf object by setting neighbourhood radius of 0.01 and 30 points as the minimum for a cluster. Then, a 3D mesh of the leaf was generated after estimating normals and ensuring that the normals were consistently orientated. Point Cloud Poisson algorithm was used for mesh generation with a depth value of 9 to preserve details. The normal estimation and their consistent orientation are essential for the poisson surface reconstruction algorithm because this surface mesh generation is based on the gradients and positions of the points, meaning that the mesh should already have appropriately aligned normals. The poisson reconstruction algorithm also returns densities of vertices in addition to the triangle mesh. They were used to generate a mask (densities < mean of densities) and the mask was used to remove low density vertices. Duplicate triangles were also removed for an accurate area estimation. After this point, a mesh was obtained with too many triangles that overestimated the leaf area. To resolve this issue, quadric decimation was used to simplify the triangle mesh of the leaf to 100 triangles, which resulted in a uniform triangle mesh. Furthermore, three iterations of Laplacian smoothing were used to ensure smooth transitions between high-density and low-density areas. The Laplacian filter moves each vertex towards the centroid of its neighbouring vertices to generate a smooth mesh by reducing sharp features and minimising noise. Finally, the area of the triangle mesh was calculated. The main steps of the image processing-based leaf area estimation workflow are shown in Figure~\ref{fig:image-processing based area} with sample images of the main steps. While optimising the parameters of the algorithms discussed above to obtain a better leaf area, we experienced the requirement of image-specific fine-tuning to obtain a better leaf area. The parameter values mentioned are obtained after trying some samples. Due to this complexity, a deep leaning-based method was investigated to predict the area of a single leaf.

%from here
\subsection{Investigation of Deep Learnining-based Area Estimation }\label{sec:method_n1}

Leaf instance segmentation is an essential and the first step for leaf-level area estimation in multi-object images. Mask R-CNN and YoLo are the emerging and competing two models in state-of-the-art instance segmentation, as mentioned in the Introduction section. The area estimation precession of a segmented object depends on the segmentation accuracy and according to the testing result metrics on COCO dataset~\citep{lin2015microsoftcococommonobjects}, Mask R-CNN provides a better accuracy below 0.75 IOU while YoLo is much faster. Therefore, Mask R-CNN~\citep{he2018maskrcnn} and Detectron2~\citep{Detectron2018} which is a framework with Mask R-CNN implementation were used to develop and test an instance segmentation and area estimation model. The two datasets explained in~\ref{subsec:data_collection}, prepared in COCO format, included RGB and depth channels, with depth aiding accurate area calculation regardless of the position of the object. Detectron2 was extended to handle four-channel RGBD inputs by modifying the data loading and augmentation processes. Separate augmentations were applied to RGB and depth images because the typical colour feature value augmentations cannot be applied to depth channel. 
A new ``Area-head'' was implemented for area estimation. 2D convolution (k = 1 layers) was performed per object instance on features (F) generated by FPN and masked using the segmentation mask to train weights W, then group normalisation and an ReLU were employed to generate features (Equation~\ref{eq:feature_transform}). After final convolution layer (n), another convolution layer was used to predict per pixel area, ReLU activation was used and reduce layer (sum) was used to generate the final object area (Equation~\ref{eq:combined_area}). 
The mask head was adjusted to output predicted masks to use in the Area-head while training in addition to losses output in original detectron2 implementation. High-level network architecture of this extended Mask R-CNN (N1) is shown in Figure~\ref{sec:sla-rcnn_n1} and conducted experiments are listed in the first four rows of Table~\ref{tab:experiments}. Selected options from each experiment are bolted and were used in next experiments.

\begin{equation}
\small{
F^{(k+1)} = \text{Activation}\left(\text{Norm}\left(\text{Conv2d}\left(F^{(k)}, W^{(k)}\right)\right)\right)\\}
\label{eq:feature_transform}
\end{equation}

\begin{equation}
\small{
A = \sum_{x, y} \text{Activation}\left(\text{Conv2d}\left(F^{(n)}, W^{\text{pred}}\right)\right)}
\label{eq:combined_area}
\end{equation}

% Figure~\ref{fig:sla-rcnn-N2}.

\begin{table}[h]
  \centering
  % \small{
  \begin{tabular*}{\linewidth}
  % {ccc}
  {@{\extracolsep{\fill}}p{0.17\linewidth}p{0.37\linewidth}p{0.44\linewidth}@{}}
    \hline
    Experiment & Param/technique & Options tried \\ \hline
    N1-1 & optimizer & SGD, \textbf{ADAM} \\ 
    N1-2 & normalisation & batch, \textbf{group} \\ 
    N1-3 & mixing of two datasets & Sequential, \textbf{mixed batch} \\ 
    N1-4 & freezing backbone & 2 layers, \textbf{0 layers} \\ 
    N2-0 & batch size & \textbf{2}, 4, 6, 8 \\ 
    N2-1 & AH-loss function & MSE, Huber, \textbf{L1} \\ 
    N2-2 & AH-Activation function &  ReLU, \textbf{Leaky ReLU} \\
    N2-3 & \#CNN layers &  \textbf{1}, 2, 4 \\ 
    N2-4 & AH-input channels & D, \textbf{DG} \\ \hline
  \end{tabular*}%}
  \caption{Experiments conducted in developing DLA-RCNN}
  \label{tab:experiments}
\end{table}

\subsection{DLA-RCNN: Deep Learning-based Direct Area Estimation}\label{sec:sla-rcnn_n1}

\begin{figure*}[t]
  \centering
   \includegraphics[width=0.6\linewidth]{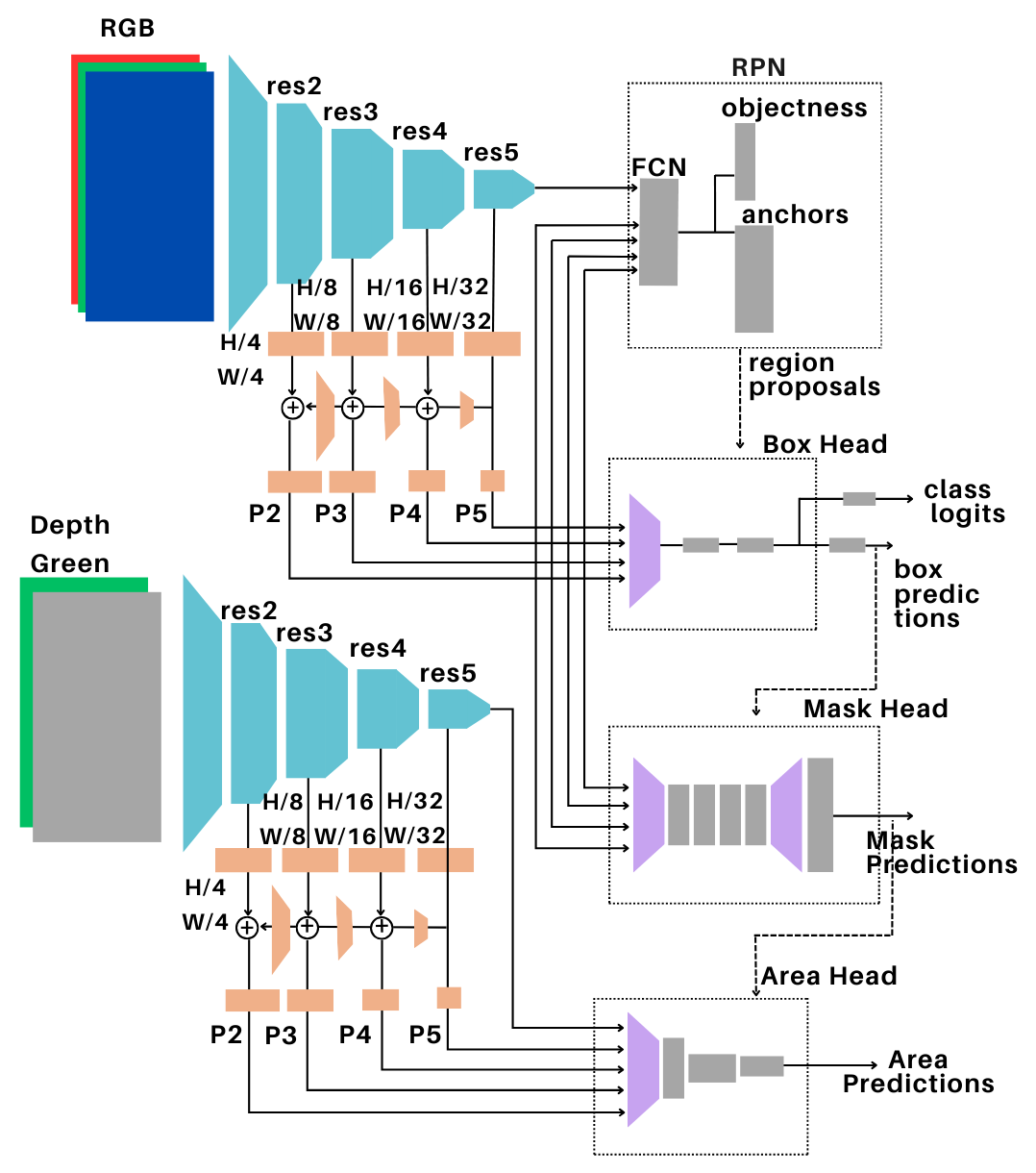}

   \caption{DLA-RCNN Network Architecture (N2) for leaf segmentation and area estimation using RGBD images}
   \label{fig:sla-rcnn-N2}
\end{figure*}

We introduced a dual backbone architecture (N2) to extract the features separately for the area estimation while leaving the segmentation with RGB features (Figure~\ref{fig:sla-rcnn-N2}). Experiments are assisted with an agile approach to investigate a better loss function, activation function, the number of CNN layers, and input channels in the area head. Details hyperparameters are shown in last five rows of Table~\ref{tab:experiments}. N2 experiments were continued with mixed batches as positivity was observed in experiment EN1-3. Different batch sizes were tried; (EN2-1) even a lower batch size should be fine because of group normalisation. Different loss functions were tried to account the noisy nature of the depth images, with the L1 loss chosen for its ability to handle outliers and the Huber loss providing a balance between L1 and L2. The LeakyReLU activation function was tested as it allows neurones to remain active even for negative values, potentially improving learning. In addition, the number of convolution layers in the area head (EN-4) and the use of the green channel in combination with depth in the second backbone were also tested. The green channel was selected because of its strong correlation with leaf reflection, which could help reduce edge noise in the depth images.

In implementing this two-backbone design in Detectron2, several modifications were required. A custom class for ResnetFPNBackbone was created to support the dual backbones. During the initialisation of the class, the first three channels (RGB) were used to build the typical backbone, and the fourth channel (depth) was used to build the second backbone. The forward function was also altered to process the fourth channel (depth) through the second backbone. Rather than returning a single RGB feature dictionary, both RGB and depth features were returned as separate dictionaries. Additionally, the Region Proposal Network (RPN) was modified to use only the RGB features from the output of the first backbone, while the ROI heads were updated to pass RGB features to the original network heads and depth features to the area head.

The performance of the model was evaluated using two metrics: the F1 score (F1$_{90}$), which measures the segmentation precision based on the intersection area over 90\% IoA, and R-squared (R$^2_{90}$), which evaluates the area estimation precision for the same intersection. After selecting the best hyperparameters for the segmentation and area estimation model, a 5-fold cross-validation was performed to verify its generalisability. The final model, showing average performance across these tests, was then used for further evaluation on unseen single-leaf and plant testing data. Then the area estimation result was compared with the typical image processing-based method (explained in Subsection~\ref{sec:image_processing}) using the same segmentation masks that were used in the deep learning-based method. In detail, correlation between ground truth area and two methods, and error distribution of area estimations/calculations over distance from camera to objects were checked.

\begin{figure}[t]
  \centering
   \includegraphics[width=1.0\linewidth]{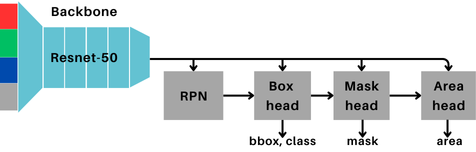}

   \caption{Highlevel architecture of trialed extended Mask R-CNN (N1) }
   \label{fig:exps-N2}
\end{figure}

\section{Results and Discussion}\label{sec:resndis}

\subsection{Model development under two network architectures and Cross-Validation}\label{sec:train_cros_v} 

\begin{figure*}[h]
  \centering
  \begin{minipage}{0.5\textwidth}
    \includegraphics[width=\linewidth]{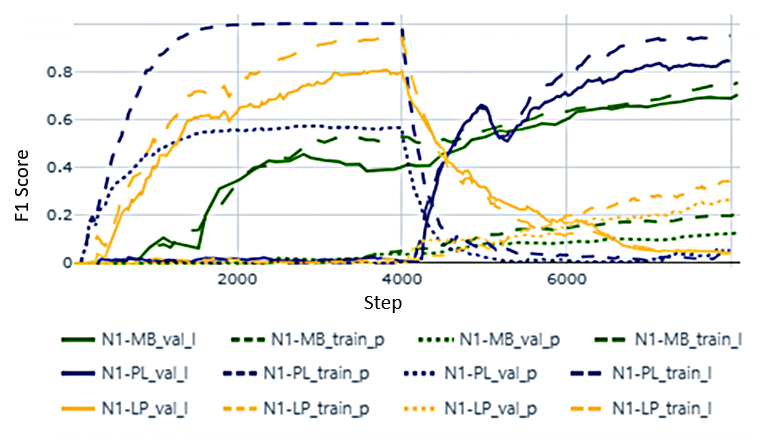}
    \subcaption{N1 segmentation}
  \end{minipage}\hfill
  \begin{minipage}{0.5\textwidth}
    \includegraphics[width=\linewidth]{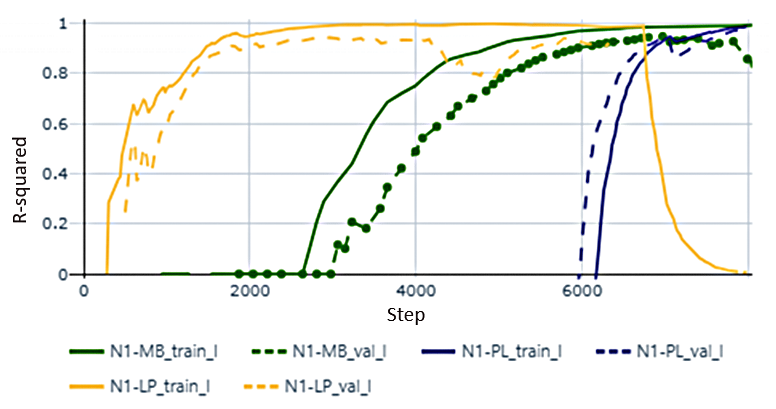}
    \subcaption{N1 area}
  \end{minipage}\hfill
  \begin{minipage}{0.5\textwidth}
    \includegraphics[width=\linewidth]{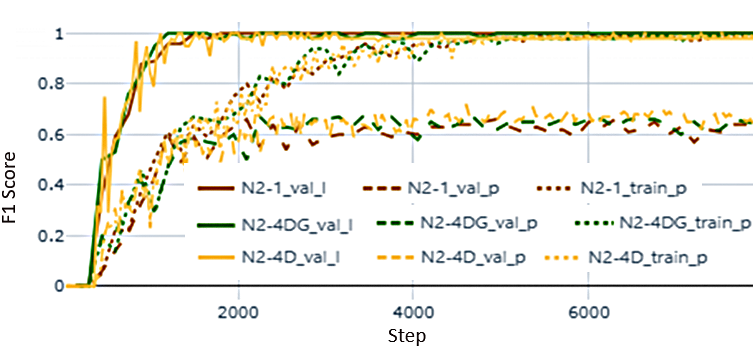}
    \subcaption{N2 segmentation}
  \end{minipage}
  % \vspace{0.5cm}
  \begin{minipage}{0.49\textwidth}
    \includegraphics[width=\linewidth]{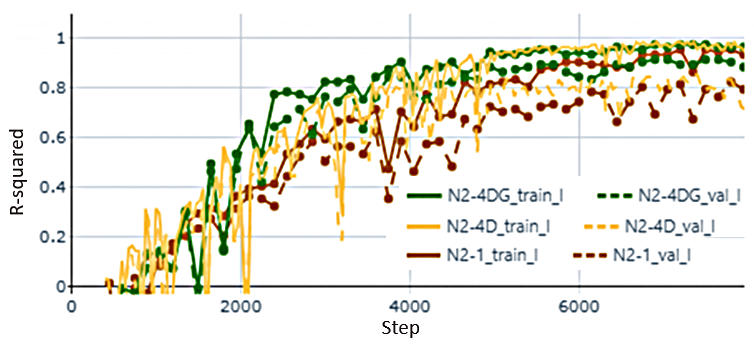}
    \subcaption{N2 area}
  \end{minipage}

  \caption{Comparison of segmentation and area estimation results under 0.9 IoA for two Network Architectures. Legend format: {network}-(experiment/parameter)\_(train/val)\_(plant data/leaf data). Notations: MB: mixed batch, PL/LP:sequencial training on leaf and plant images, N2-1/4: 1$^{st}$ or 4$^{th}$ exp with N2, D/DG: depth/depth and green input for area estimation.}
  \label{fig:model_training}
\end{figure*}

The network architecture-1 performed well for detached leaves in both segmentation and area estimation with ADAM optimiser due to its adaptability to noisy gradients and group normalisation that work independent of batch size. This can be visually noticed at Figure~\ref{fig:model_training} (a, b) N1-LP\_train/val\_l) up to 4000 iterations where the model is trained on detached single-leaf data. As shown in those two sub figures, sequential training always reduced the metric values when trained with a second dataset (yellow and blue in (a)) due to the too different backgrounds around annotated leaves in the two datasets. The mixed batch which used samples from both datasets in a single batch could not learn segmentation on plant leaves (green dash and dot lines in (a)) with this network architecture. For the models trained well for segmentation on single leaves, showed positive results for area estimation. 

Network architecture-2 performed relatively well for mixing two data sets in segmentation with mixed batch approach even though an overfitting is noted as shown in Figure~\ref{fig:model_training} (c). It was superior in segmentation for detached-leaf images while F1$_{90}$ was limiting to 0.65 for plant images. All of the model development experiments with network architecture-2 showed this trend. Training of the Network architecture-1 only with plant data (first 400 iterations of Figure~\ref{fig:model_training} (c) N1-PL\_train\_p, N1-PL\_val\_p) also showed this overfitting. All experiments with network architecture-2 for optimisation of area head showed similar patterns of training. However, the mixed batch and area head with L1 loss, Leaky ReLU, single CNN layer performed slightly better.
Using both Green and depth channels in the second backbone for area estimation performed better witnessing the support of handling depth errors using colour features in area prediction. See R$^2_{90}$ with validation data (Figure~\ref{fig:model_training} (d) N2-4DG\_val\_l) that shows values closer to R$^2_{90}$ with training data and higher values compared to using only the depth channel in the second backbone. 
Overall, this two-backbone design with RGB channels input for segmentation and depth and green channels input for area estimation was found to be a good design for mixing two datasets in training the leaf area estimation model. The selected hyperparameters for the final model are shown in bold text in Table~\ref{tab:experiments}.

A 5-fold cross-validation was conducted using the best model configuration and results are shown in Table~\ref{tab:cross_val}. All five combinations of the dataset show similar results without large deviations. This is evident that the standard deviations are close to zero for all the metrics obtained with five dataset splits. Therefore, the model can be considered unbiased with respect to the data splits. The averages of the metrics were also calculated across five folds to select a model for further testing. The model developed with dataset split-1 (set 1 in Table~\ref{tab:cross_val}) was closer to the average metric values and was used for further unseen testing.

\begin{table*}
  \centering{
  % \small{
  \begin{tabular*}{\linewidth}
  % {@{}ccccccc@{}}
  {@{\extracolsep{\fill}}p{0.08\linewidth}p{0.12\linewidth}p{0.12\linewidth}p{0.13\linewidth}p{0.13\linewidth}p{0.07\linewidth}p{0.07\linewidth}@{}}
    \hline
    Data & Training Size (leaf/plant) & Validation Size (leaf/plant) & Segm F1$_{90}$ (leaf/plant) & Avg IoA$_{90}$  (leaf/plant) & Avg Conf (leaf/plant) & Area R2$_{90}$ \\ \hline
    Set 1 & 220/1059 & 58/268 & 1.0/0.64 & 0.98/0.95 & 1/1 & 0.88 \\
    Set 2 & 220/1052 & 58/275 & 0.98/0.7 & 0.98/0.96 & 1/1 & 0.92 \\
    Set 3 & 221/1066 & 57/261 & 1.0/0.68 & 0.97/0.96 & 1/1 & 0.7 \\
    Set 4 & 224/1071 & 54/256 & 1.0/0.55 & 0.97/0.95 & 1/1 & 0.89 \\
    Set 5 & 227/1060 & 51/267 & 1.0/0.59 & 0.98/0.96 & 1/1 & 0.91 \\ \hline
    Avg & 222/1062 & 56/265 & 1.0/0.63 & 0.98/0.96 & 1/1 & 0.86 \\
    Std & 3.05/7.23 & 3.05/7.23 & 0.01/0.06 & 0.01/0.01 & 0/0 & 0.09 \\ \hline
  \end{tabular*}}%}
  \caption{5-fold cross-validation of the model developed with selected hyper-parameters (N2-architecture)}
  \label{tab:cross_val}
\end{table*}

\begin{table*}%[h]
  \centering{
  % \small{
  \begin{tabular*}{\linewidth}
  % {|c|c|c|c|c|c|c|c|c|}
{p{0.12\linewidth}p{0.08\linewidth}p{0.08\linewidth}p{0.08\linewidth}p{0.08\linewidth}p{0.08\linewidth}p{0.08\linewidth}p{0.08\linewidth}p{0.08\linewidth}}
    \hline
    Image Data & & & Validation & & & Testing & & \\ \hline
     & Seg F1$_{90}$ & Avg IoA$_{90}$ & Avg Conf & Area R2$_{90}$ & Seg F1$_{90}$ & Avg IoA$_{90}$ & Avg Conf & Area R2$_{90}$ \\ \hline
    Detached leaf & 1.0  & 0.98 & 1.0 & 0.9 & 1.0 & 0.96 & 1.0 & 0.81 \\ \hline
    On-plant leaves & 0.65 & 0.96 & 1.0 & - & 0.59 & 0.94 & 0.99 & 0.57 \\ \hline
  \end{tabular*}}%}
  \caption{Segmentation (F1-score) and area estimation (R$^2$) results of validation and unseen testing data}
  \label{tab:test_results}
\end{table*}

\subsection{Segmentation and Area Estimation Results }\label{sec:train_cros_v}

\begin{figure*}[h]
  \centering
  \includegraphics[width=0.8\textwidth]{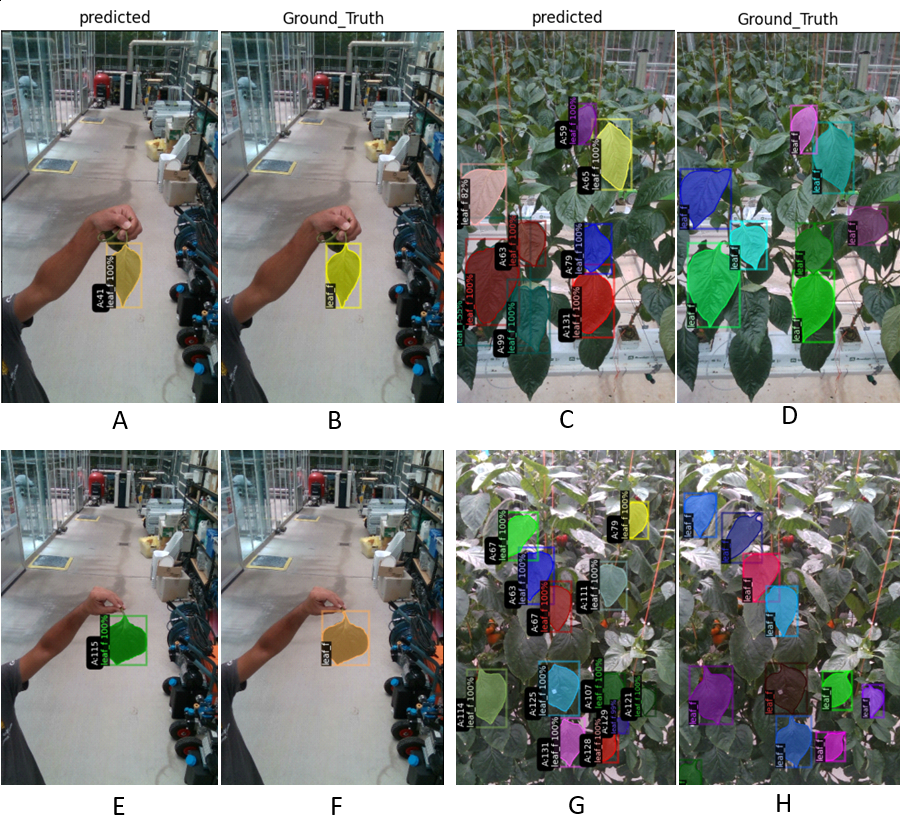}
  \caption{Visual samples of segmentation and area estimation test cases. A,B: single leaf validation; C,D: On plant validation; E, F: Single leaf testing; G, H: On plant testing }
  \label{fig:test_samples}
\end{figure*}

The selected DLA-RCNN model was evaluated on validation plant data, validation leaf data, unseen leaf data, and unseen plant data. Table~\ref{tab:test_results} presents the F1 scores, R-squared values (area estimation), average IoA values, and confidence scores. For validation and test leaf data, predictions with over 90\% IoA and confidence scores achieved 1.0 F1 scores. The R-squared for unseen leaf data was 0.81, compared to 0.9 for validation data, indicating a 9\% accuracy drop, which is acceptable. For attached-leaf data, the segmentation F1 score was 0.65 for the validation dataset. Similar results were observed in model development experiments with attached-leaf data alone and mixed datasets, indicating a segmentation overfitting. However, predictions showed most leaves were correctly segmented, with some unannotated leaves (but can be considered fully visible because annotations were determined by eye balling and there was no straight line of separation) detected as fully visible, contributing to false positives and lower F1 scores. This can be visually seen in the sample images in Figure~\ref{fig:test_samples}. The unseen plant data achieved a 0.59 F1 score, 6\% lower than the validation data, and a 0.57 R-squared for the area estimation accuracy, 24\% lower than the unseen leaf data.

\begin{figure*}[h]
  \centering
  \begin{minipage}{0.4\textwidth}
    \includegraphics[width=\linewidth]{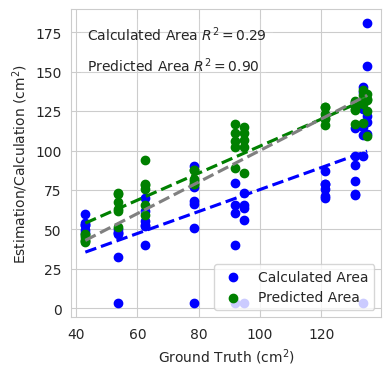}
    \subcaption{Validation leaf data}
  \end{minipage}\hfill
  \begin{minipage}{0.4\textwidth}
    \includegraphics[width=\linewidth]{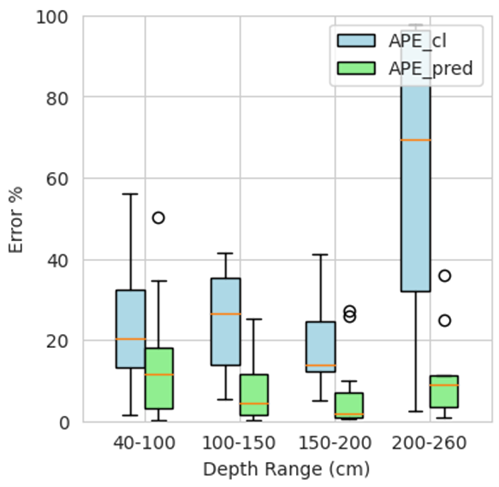}
    \subcaption{Testing leaf data}
  \end{minipage}\hfill
  \begin{minipage}{0.4\textwidth}
    \includegraphics[width=\linewidth]{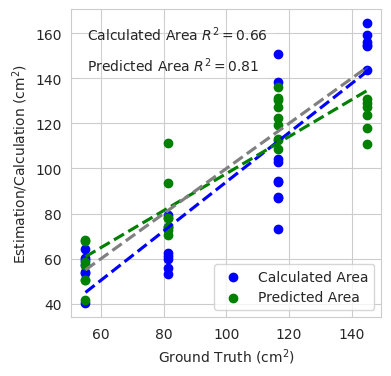}
    \subcaption{Testing plant data}
  \end{minipage}
  % \vspace{0.5cm}
  \begin{minipage}{0.4\textwidth}
    \includegraphics[width=\linewidth]{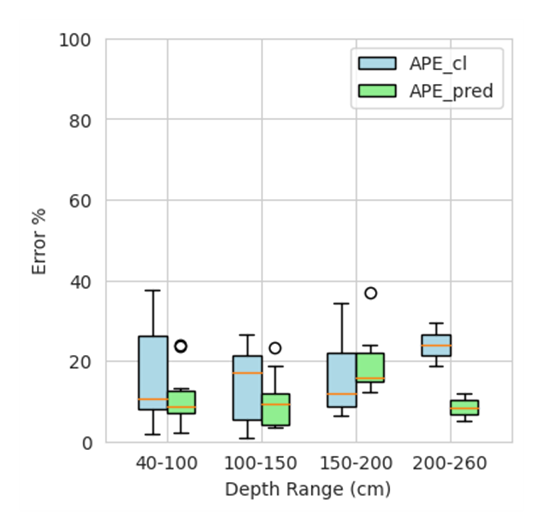}
    \subcaption{Validation leaf data}
  \end{minipage}\hfill
  \begin{minipage}{0.4\textwidth}
    \includegraphics[width=\linewidth]{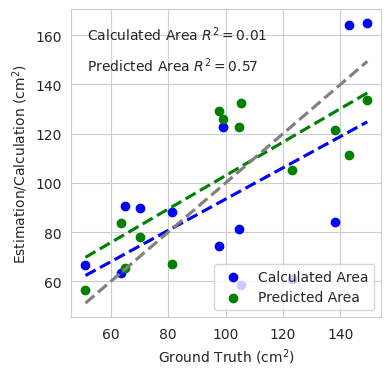}
    \subcaption{Testing leaf data}
  \end{minipage}\hfill
  \begin{minipage}{0.4\textwidth}
    \includegraphics[width=\linewidth]{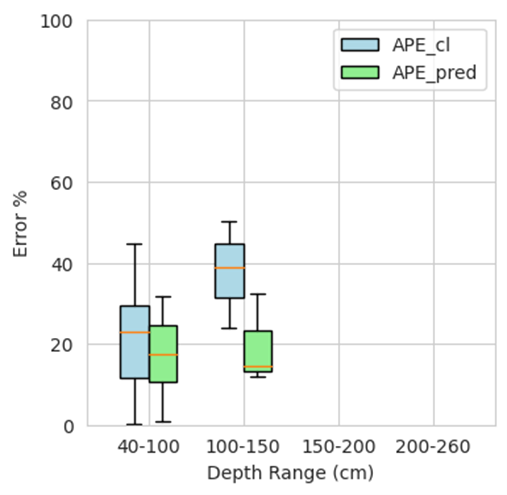}
    \subcaption{Testing plant data}
  \end{minipage}
  \caption{Correlations of area estimations over ground truths (a, b, and c) and error distributions of area estimation over distance bins (d, e, and f).}
  \label{fig:result_plots}
\end{figure*}

The comparison between DLA-RCNN and image processing-based area calculations, shown in Figure~\ref{fig:result_plots} and error statistics in Table~\ref{tab:DL_vs_IP}, reveals that the deep learning approach is better. For validation data, the deep learning method achieved an R-squared value of 0.90, significantly outperforming the image processing method, which had an R-squared of 0.29. The error distribution for DLA-RCNN's predictions was consistent across the leaves at different distances, while image processing results showed substantial variability. This pattern was also observed in the unseen detached-leaf test data, where DLA-RCNN maintained an R-squared of 0.81, compared to 0.66 for image processing. For unseen plant leaves, the performance was more challenging. The segmentation of attached leaves on plants achieved an R-squared of 0.57 for deep learning predictions and only 0.01 for image processing. In general, depth noise from complex backgrounds reduced overall performance. As illustrated in Figure~\ref{fig:data} depth maps for mid-to-lower plant leaves were noisier compared to those for the top leaves, contributing to higher errors.
Although both methods showed fluctuated errors with distance across validation and testing data sets, image processing-based method showed higher error trend after 2m. We suspect that the deep learning model has generalised to handle noisy depth images captured with more distance and has compromised accuracy of images captured closer.  
% The mean error for deep learning for leaves with 1 m distance was 7.35\% with a standard deviation of 18.10\%, while image processing errors were showing similar error distribution. Beyond 1 m, deep learning provided better accuracy while image processing showing a lower accuracy. 

Absolute percentage error (APE) of area predictions were also calculated to investigate the reliability of the methods as shown in Table~\ref{tab:DL_vs_IP}. DLA-RCNN shows relatively lower values for mean, median, and std of APE compared to those of the image preprocessing-based calculations. Better medians and a bit higher stds can be observed. The noisy depth maps for leaves in mid and lower areas of plant images (\ref{fig:data}) could have contributed to higher means and stds. 

Overall, DLA-RCNN showed high accuracy in segmenting detached single leaves which shows reliable segmentation for busy plant images. The proposed model handles the complexities better than the traditional image processing-based method in area estimation and shows more room for improvements.

\begin{table*}%[htb]
  \centering{
  \small{
  \begin{tabular*}{\linewidth}
  {p{0.15\linewidth}p{0.07\linewidth}p{0.07\linewidth}p{0.07\linewidth}p{0.07\linewidth}p{0.07\linewidth}p{0.07\linewidth}p{0.07\linewidth}p{0.07\linewidth}}
    \hline
    Image Data & \multicolumn{4}{c}{APE of DLA-RCNN}  & \multicolumn{4}{c}{APE of image processing}  \\ 
    & R$^2$ & Mean & Std & Med & R$^2$ & Mean & Std & Med \\ \hline
    Validation leaf & 0.9 & 9.71 & 10.48 & 5.1 & 0.29 & 29.41 & 24.05 & 24.15 \\ 
    Test leaf & 0.81 & 12.15 & 7.95 & 10.89 & 0.66 & 16.01 & 10.52 & 13.47 \\ 
    Test plant & 0.57 & 17.9 & 8.98 & 17.42 & 0.01 & 25.79 & 14.36 & 24.02 \\ \hline
  \end{tabular*}}}
  \caption{ Distribution of absolute percentage error for both calculated area (traditional image processing-based method) and predicted area (DLA-RCNN)}
  \label{tab:DL_vs_IP}
\end{table*}

\subsection{Practical Usability, Limitations, and Future Directions }\label{sec:future}
% Practical Usability
This study demonstrates a deep learning-based methodology that effectively segments fully visible leaves, both detached and on plants. The deep learning method shows better accuracy and reliability for segmentation and leaf area estimation.
% when the camera is positioned within 1.5 meter. 
This approach can be applied to estimate the area of fully visible plant leaves where the camera is moving or fixed. Specially for estimating area of fully visible leaves in top areas of tall plants in glasshouses when top-angled moving camera setups are used to estimate plant heights\citep{jayasuriya_machine_2024}. Compared to recent work using ideal camera positions, manual selection of segmented leaves, and image processing~\cite{pyau_portable_2021}, our method demonstrated a precision of 0.81 R$^2_{90}$  for separated leaves, outperforming traditional image-based methods (0.66). For on-plant leaves in noisy, top-angle views, R$^2_{90}$ was 0.57, highlighting the need for further research on focused, noise-reduced data.
% Additionally, the method can be detect leaf-level stress and diseases by integrating anomaly detection, offering valuable insights for plant health monitoring.

% Limitations
The model is tested with a top-angle view camera only for the leaves at top to middle of the plant that are within 150 cm range. The middle to lower leaves which are away will have poor visibility and higher depth noise will result in higher error. Any depth image with highly distorted depth data will also leads to lower accuracies. A relatively good accuracy can be achieved when leaves are within 1.5 m distance from camera and 2 m range can also be considered. The leaves which have higher angles relative to the camera plane also will increase the depth noise resulting in lower accuracy. The model was trained only on capsicum leaves and plants and it needs transfer learning using other datasets to work on other plants.

% Future Directions
Future improvements could involve expanding the model to recognise additional leaf classes, including partially visible, angled, and backside leaves. This expansion would enhance the segmentation accuracy and also can be extended towards total leaf area estimation using number of leaves and average leaf size. A much better total leaf area can be estimated with in-painting occluded leaves. Capturing ground truth data for leaves on plants would improve training robustness and would be able to handle more depth noises occurring due to complex background resulting in a better area estimation accuracy. Using a vertically moving camera with an advanced hardware system can capture better images of mid and lower parts of the plants to get more accurate area estimations. Additionally, comparing the resource and time efficiency of deep learning versus traditional image processing methods will inform practical deployment in commercial settings.

\section{Conclusion}\label{sec:conclusion}

The existing image processing-based single leaf area estimations depend on unsupervised clustering-based segmentation which may need manual object selection and 3D reconstruction. This work investigated deep learning-based direct leaf area estimation for fully visible leaves for in situ conditions. The method shows mixing two datasets for convenient ground truth collection for model training with detached leaves and applying for attached leaves. Further, an agile way of hyperparameter tuning was demonstrated. DLA-RCNN with two backbone architectures using
green and depth features for area estimation outperformed single backbone architecture and traditional image processing-based method. DLA-RCNN showed F1$_{90}$ of 1.0 with high confidence for off the plant leaves while it was 0.59 on attached leaves. DLA-RCNN claimed 0.81 R$^2_{90}$ on detached leaves and 0.57 on attached leaves with relatively good APE, while it was showing varied correlations with image processing-based method on validation and testing datasets.  In conclusion, deep learning methodology showed positive and more reliable results and further improvements are needed for in situ use.
%%%%%%%%% REFERENCES

% \appendix

% ------------------------------------------------------------------------

\printcredits

\textbf{Acknowledgements:} We acknowledge the research director at NVPCC Distinguished Professor David Tissue, researcher Dr. Sachin Chavan, Crop Manager Wei Liang, Facility Manager Goran Lopaticki, Norbert Klause and others who support our work at NVPCC.

\textbf{Funding:} This research is jointly funded by Western Sydney University and the Future Food Systems Cooperative Research Centre.

\bibliographystyle{cas-model2-names}

\bibliography{cas-refs}

\begin{thebibliography}{29}
\expandafter\ifx\csname natexlab\endcsname\relax\def\natexlab#1{#1}\fi
\providecommand{\url}[1]{\texttt{#1}}
\providecommand{\href}[2]{#2}
\providecommand{\path}[1]{#1}
\providecommand{\DOIprefix}{doi:}
\providecommand{\ArXivprefix}{arXiv:}
\providecommand{\URLprefix}{URL: }
\providecommand{\Pubmedprefix}{pmid:}
\providecommand{\doi}[1]{\href{http://dx.doi.org/#1}{\path{#1}}}
\providecommand{\Pubmed}[1]{\href{pmid:#1}{\path{#1}}}
\providecommand{\bibinfo}[2]{#2}
\ifx\xfnm\relax \def\xfnm[#1]{\unskip,\space#1}\fi
%Type = Article
\bibitem[{Baret et~al.(2010)Baret, de~Solan, Lopez-Lozano, Ma and Weiss}]{baret_gai_2010}
\bibinfo{author}{Baret, F.}, \bibinfo{author}{de~Solan, B.}, \bibinfo{author}{Lopez-Lozano, R.}, \bibinfo{author}{Ma, K.}, \bibinfo{author}{Weiss, M.}, \bibinfo{year}{2010}.
\newblock \bibinfo{title}{{GAI} estimates of row crops from downward looking digital photos taken perpendicular to rows at 57.5° zenith angle: {Theoretical} considerations based on {3D} architecture models and application to wheat crops}.
\newblock \bibinfo{journal}{Agricultural and Forest Meteorology} \bibinfo{volume}{150}, \bibinfo{pages}{1393--1401}.
\newblock \URLprefix \url{https://www.sciencedirect.com/science/article/pii/S0168192310001206}, \DOIprefix\doi{10.1016/j.agrformet.2010.04.011}.
%Type = Misc
\bibitem[{{CVAT.ai Corporation}(2023)}]{CVAT_2023}
\bibinfo{author}{{CVAT.ai Corporation}}, \bibinfo{year}{2023}.
\newblock \bibinfo{title}{{Computer Vision Annotation Tool (CVAT)}}.
\newblock \URLprefix \url{https://github.com/cvat-ai/cvat}.
%Type = Article
\bibitem[{DaMatta(2004)}]{damatta_exploring_2004}
\bibinfo{author}{DaMatta, F.M.}, \bibinfo{year}{2004}.
\newblock \bibinfo{title}{Exploring drought tolerance in coffee: a physiological approach with some insights for plant breeding}.
\newblock \bibinfo{journal}{Brazilian Journal of Plant Physiology} \bibinfo{volume}{16}, \bibinfo{pages}{1--6}.
\newblock \URLprefix \url{https://www.scielo.br/j/bjpp/a/CxBgpsX6cpbrtNmHSFmqJXK/?lang=en}, \DOIprefix\doi{10.1590/S1677-04202004000100001}. \bibinfo{note}{publisher: Brazilian Journal of Plant Physiology}.
%Type = Article
\bibitem[{De~Bock et~al.(2023)De~Bock, Belmans, Vanlanduit, Blom, Alvarado-Alvarado and Audenaert}]{de_bock_review_2023}
\bibinfo{author}{De~Bock, A.}, \bibinfo{author}{Belmans, B.}, \bibinfo{author}{Vanlanduit, S.}, \bibinfo{author}{Blom, J.}, \bibinfo{author}{Alvarado-Alvarado, A.A.}, \bibinfo{author}{Audenaert, A.}, \bibinfo{year}{2023}.
\newblock \bibinfo{title}{A review on the leaf area index ({LAI}) in vertical greening systems}.
\newblock \bibinfo{journal}{Building and Environment} \bibinfo{volume}{229}, \bibinfo{pages}{109926}.
\newblock \URLprefix \url{https://www.sciencedirect.com/science/article/pii/S0360132322011568}, \DOIprefix\doi{10.1016/j.buildenv.2022.109926}.
%Type = Inproceedings
\bibitem[{Ester et~al.(1996)Ester, Kriegel, Sander, Xu et~al.}]{ester1996dbscan}
\bibinfo{author}{Ester, M.}, \bibinfo{author}{Kriegel, H.P.}, \bibinfo{author}{Sander, J.}, \bibinfo{author}{Xu, X.}, et~al., \bibinfo{year}{1996}.
\newblock \bibinfo{title}{A density-based algorithm for discovering clusters in large spatial databases with noise}, in: \bibinfo{booktitle}{kdd}, pp. \bibinfo{pages}{226--231}.
%Type = Misc
\bibitem[{Girshick et~al.(2018)Girshick, Radosavovic, Gkioxari, Doll\'{a}r and He}]{Detectron2018}
\bibinfo{author}{Girshick, R.}, \bibinfo{author}{Radosavovic, I.}, \bibinfo{author}{Gkioxari, G.}, \bibinfo{author}{Doll\'{a}r, P.}, \bibinfo{author}{He, K.}, \bibinfo{year}{2018}.
\newblock \bibinfo{title}{Detectron}.
\newblock \bibinfo{howpublished}{\url{https://github.com/facebookresearch/detectron}}.
%Type = Misc
\bibitem[{He et~al.(2018)He, Gkioxari, Dollár and Girshick}]{he2018maskrcnn}
\bibinfo{author}{He, K.}, \bibinfo{author}{Gkioxari, G.}, \bibinfo{author}{Dollár, P.}, \bibinfo{author}{Girshick, R.}, \bibinfo{year}{2018}.
\newblock \bibinfo{title}{Mask r-cnn}.
\newblock \URLprefix \url{https://arxiv.org/abs/1703.06870}, \href{http://arxiv.org/abs/1703.06870}{\tt arXiv:1703.06870}.
%Type = Article
\bibitem[{Hu et~al.(2023)Hu, Tang, Shi and Qian}]{hu_detection_2023_yolo_mrcnn}
\bibinfo{author}{Hu, H.}, \bibinfo{author}{Tang, C.}, \bibinfo{author}{Shi, C.}, \bibinfo{author}{Qian, Y.}, \bibinfo{year}{2023}.
\newblock \bibinfo{title}{Detection of residual feed in aquaculture using {YOLO} and {Mask} {RCNN}}.
\newblock \bibinfo{journal}{Aquacultural Engineering} \bibinfo{volume}{100}, \bibinfo{pages}{102304}.
\newblock \URLprefix \url{https://www.sciencedirect.com/science/article/pii/S0144860922000802}, \DOIprefix\doi{10.1016/j.aquaeng.2022.102304}.
%Type = Article
\bibitem[{Hu et~al.(2020)Hu, Lu, Meng, Li, Cong, Ren, Sharkey and Lu}]{hu_reduction_2020}
\bibinfo{author}{Hu, W.}, \bibinfo{author}{Lu, Z.}, \bibinfo{author}{Meng, F.}, \bibinfo{author}{Li, X.}, \bibinfo{author}{Cong, R.}, \bibinfo{author}{Ren, T.}, \bibinfo{author}{Sharkey, T.D.}, \bibinfo{author}{Lu, J.}, \bibinfo{year}{2020}.
\newblock \bibinfo{title}{The reduction in leaf area precedes that in photosynthesis under potassium deficiency: the importance of leaf anatomy}.
\newblock \bibinfo{journal}{New Phytologist} \bibinfo{volume}{227}, \bibinfo{pages}{1749--1763}.
\newblock \URLprefix \url{https://onlinelibrary.wiley.com/doi/abs/10.1111/nph.16644}, \DOIprefix\doi{10.1111/nph.16644}. \bibinfo{note}{\_eprint: https://onlinelibrary.wiley.com/doi/pdf/10.1111/nph.16644}.
%Type = Misc
\bibitem[{Inc(a)}]{cid_inc_ci-202_nodate}
\bibinfo{author}{Inc, L.C.}, a.
\newblock \bibinfo{title}{{CI}-202 {Portable} {Laser} {Leaf} {Area} {Meter}}.
\newblock \URLprefix \url{https://cid-inc.com/plant-science-tools/leaf-area-measurement/ci-202-portable-laser-leaf-area-meter/}.
%Type = Misc
\bibitem[{Inc(b)}]{li-cor_inc_li-3100c_nodate}
\bibinfo{author}{Inc, L.C.}, b.
\newblock \bibinfo{title}{{LI}-{3100C} {Area} {Meter}}.
\newblock \URLprefix \url{https://www.licor.com/env/products/leaf-area/LI-3100C/}.
%Type = Manual
\bibitem[{Intel-RealSenseTM(2023)}]{intel-realsense-d400-datasheet}
\bibinfo{author}{Intel-RealSenseTM}, \bibinfo{year}{2023}.
\newblock \bibinfo{title}{Intel RealSense D400 Series Datasheet}.
\newblock \URLprefix \url{https://www.intelrealsense.com/wp-content/uploads/2023/03/Intel-RealSense-D400-Series-Datasheet-March-2023.pdf}. \bibinfo{note}{revision 015, Document Number: 337029-013}.
%Type = Article
\bibitem[{Jayasuriya et~al.(2024a)Jayasuriya, Ghannoum, Hu, Klause, Liang and Guo}]{jayasuriya_pc4c_capsi_2024}
\bibinfo{author}{Jayasuriya, N.}, \bibinfo{author}{Ghannoum, O.}, \bibinfo{author}{Hu, W.}, \bibinfo{author}{Klause, N.}, \bibinfo{author}{Liang, W.}, \bibinfo{author}{Guo, Y.}, \bibinfo{year}{2024}a.
\newblock \bibinfo{title}{{PC4C}\_capsi: {Image} data of capsicum plant growth in protected horticulture}.
\newblock \bibinfo{journal}{Data in Brief} \bibinfo{volume}{55}, \bibinfo{pages}{110735}.
\newblock \URLprefix \url{https://www.sciencedirect.com/science/article/pii/S2352340924007029}, \DOIprefix\doi{10.1016/j.dib.2024.110735}.
%Type = Article
\bibitem[{Jayasuriya et~al.(2024b)Jayasuriya, Guo, Hu and Ghannoum}]{jayasuriya_machine_2024}
\bibinfo{author}{Jayasuriya, N.}, \bibinfo{author}{Guo, Y.}, \bibinfo{author}{Hu, W.}, \bibinfo{author}{Ghannoum, O.}, \bibinfo{year}{2024}b.
\newblock \bibinfo{title}{Machine vision based plant height estimation for protected crop facilities}.
\newblock \bibinfo{journal}{Computers and Electronics in Agriculture} \bibinfo{volume}{218}, \bibinfo{pages}{108669}.
\newblock \URLprefix \url{https://www.sciencedirect.com/science/article/pii/S0168169924000607}, \DOIprefix\doi{10.1016/j.compag.2024.108669}.
%Type = Misc
\bibitem[{Jayasuriya et~al.(2024c)Jayasuriya, Weerasekara, Ghannoum, Guo and Hu}]{jayasuriya_spi-vstl_2024}
\bibinfo{author}{Jayasuriya, N.}, \bibinfo{author}{Weerasekara, M.}, \bibinfo{author}{Ghannoum, O.}, \bibinfo{author}{Guo, Y.}, \bibinfo{author}{Hu, W.}, \bibinfo{year}{2024}c.
\newblock \bibinfo{title}{Spi-{Vstl}: {Image} {Data} {Collection} {Platform} {Using} {Off}-the-{Shelf} {Hardware} for {Vertically} {Supported} {Crops} in {State}-of-{The}-{Art} {Glasshouses}}.
\newblock \URLprefix \url{https://papers.ssrn.com/abstract=5021295}, \DOIprefix\doi{10.2139/ssrn.5021295}.
%Type = Misc
\bibitem[{Jocher et~al.(2023)Jocher, Qiu and Chaurasia}]{Jocher_Ultralytics_YOLO_2023}
\bibinfo{author}{Jocher, G.}, \bibinfo{author}{Qiu, J.}, \bibinfo{author}{Chaurasia, A.}, \bibinfo{year}{2023}.
\newblock \bibinfo{title}{{Ultralytics YOLO}}.
\newblock \URLprefix \url{https://github.com/ultralytics/ultralytics}.
%Type = Article
\bibitem[{Khan et~al.(2016)Khan, Banday, Narayan, Khan and Bhat}]{khan_use_2016}
\bibinfo{author}{Khan, F.A.}, \bibinfo{author}{Banday, F.A.}, \bibinfo{author}{Narayan, S.}, \bibinfo{author}{Khan, F.U.}, \bibinfo{author}{Bhat, S.A.}, \bibinfo{year}{2016}.
\newblock \bibinfo{title}{Use of {Models} as {Non}-destructive {Method} for {Leaf} {Area} {Estimation} in {Horticultural} {Crops}}.
\newblock \bibinfo{journal}{IRA-International Journal of Applied Sciences (ISSN 2455-4499)} \bibinfo{volume}{4}.
\newblock \URLprefix \url{http://research-advances.org/index.php/IRAJAS/article/view/381}, \DOIprefix\doi{10.21013/jas.v4.n1.p19}.
%Type = Article
\bibitem[{Lai et~al.(2022)Lai, Mu, Li, Zou, Bian, Zhou, Hu, Li, Xie and Yan}]{lai_correcting_2022}
\bibinfo{author}{Lai, Y.}, \bibinfo{author}{Mu, X.}, \bibinfo{author}{Li, W.}, \bibinfo{author}{Zou, J.}, \bibinfo{author}{Bian, Y.}, \bibinfo{author}{Zhou, K.}, \bibinfo{author}{Hu, R.}, \bibinfo{author}{Li, L.}, \bibinfo{author}{Xie, D.}, \bibinfo{author}{Yan, G.}, \bibinfo{year}{2022}.
\newblock \bibinfo{title}{Correcting for the clumping effect in leaf area index calculations using one-dimensional fractal dimension}.
\newblock \bibinfo{journal}{Remote Sensing of Environment} \bibinfo{volume}{281}, \bibinfo{pages}{113259}.
\newblock \URLprefix \url{https://www.sciencedirect.com/science/article/pii/S0034425722003650}, \DOIprefix\doi{10.1016/j.rse.2022.113259}.
%Type = Misc
\bibitem[{Lin et~al.(2015)Lin, Maire, Belongie, Bourdev, Girshick, Hays, Perona, Ramanan, Zitnick and Dollár}]{lin2015microsoftcococommonobjects}
\bibinfo{author}{Lin, T.Y.}, \bibinfo{author}{Maire, M.}, \bibinfo{author}{Belongie, S.}, \bibinfo{author}{Bourdev, L.}, \bibinfo{author}{Girshick, R.}, \bibinfo{author}{Hays, J.}, \bibinfo{author}{Perona, P.}, \bibinfo{author}{Ramanan, D.}, \bibinfo{author}{Zitnick, C.L.}, \bibinfo{author}{Dollár, P.}, \bibinfo{year}{2015}.
\newblock \bibinfo{title}{Microsoft coco: Common objects in context}.
\newblock \URLprefix \url{https://arxiv.org/abs/1405.0312}, \href{http://arxiv.org/abs/1405.0312}{\tt arXiv:1405.0312}.
%Type = Article
\bibitem[{Niu et~al.(2018)Niu, Feng, Yang, Li, Yang, Xu and Zhao}]{niu_monitoring_2018}
\bibinfo{author}{Niu, Q.}, \bibinfo{author}{Feng, H.}, \bibinfo{author}{Yang, G.}, \bibinfo{author}{Li, C.}, \bibinfo{author}{Yang, H.}, \bibinfo{author}{Xu, B.}, \bibinfo{author}{Zhao, Y.}, \bibinfo{year}{2018}.
\newblock \bibinfo{title}{Monitoring plant height and leaf area index of maize breeding material based on {UAV} digital images}.
\newblock \bibinfo{journal}{Transactions of the Chinese Society of Agricultural Engineering} \bibinfo{volume}{34}, \bibinfo{pages}{73--82}.
%Type = Article
\bibitem[{Phoncharoen et~al.(2022)Phoncharoen, Banterng, Vorasoot, Jogloy and Theerakulpisut}]{phoncharoen_determination_2022}
\bibinfo{author}{Phoncharoen, P.}, \bibinfo{author}{Banterng, P.}, \bibinfo{author}{Vorasoot, N.}, \bibinfo{author}{Jogloy, S.}, \bibinfo{author}{Theerakulpisut, P.}, \bibinfo{year}{2022}.
\newblock \bibinfo{title}{Determination of {Cassava} {Leaf} {Area} for {Breeding} {Programs}}.
\newblock \bibinfo{journal}{Agronomy} \bibinfo{volume}{12}, \bibinfo{pages}{3013}.
\newblock \URLprefix \url{https://www.mdpi.com/2073-4395/12/12/3013}, \DOIprefix\doi{10.3390/agronomy12123013}. \bibinfo{note}{number: 12 Publisher: Multidisciplinary Digital Publishing Institute}.
%Type = Inproceedings
\bibitem[{Prasetyo et~al.(2020)Prasetyo, Suciati and Fatichah}]{fish_segment2024_yolo_mrcnn}
\bibinfo{author}{Prasetyo, E.}, \bibinfo{author}{Suciati, N.}, \bibinfo{author}{Fatichah, C.}, \bibinfo{year}{2020}.
\newblock \bibinfo{title}{A comparison of yolo and mask r-cnn for segmenting head and tail of fish}, in: \bibinfo{booktitle}{2020 4th International Conference on Informatics and Computational Sciences (ICICoS)}, pp. \bibinfo{pages}{1--6}.
\newblock \DOIprefix\doi{10.1109/ICICoS51170.2020.9299024}.
%Type = Misc
\bibitem[{Ravi et~al.(2024)Ravi, Gabeur, Hu, Hu, Ryali, Ma, Khedr, Rädle, Rolland, Gustafson, Mintun, Pan, Alwala, Carion, Wu, Girshick, Dollár and Feichtenhofer}]{ravi2024sam2segmentimages}
\bibinfo{author}{Ravi, N.}, \bibinfo{author}{Gabeur, V.}, \bibinfo{author}{Hu, Y.T.}, \bibinfo{author}{Hu, R.}, \bibinfo{author}{Ryali, C.}, \bibinfo{author}{Ma, T.}, \bibinfo{author}{Khedr, H.}, \bibinfo{author}{Rädle, R.}, \bibinfo{author}{Rolland, C.}, \bibinfo{author}{Gustafson, L.}, \bibinfo{author}{Mintun, E.}, \bibinfo{author}{Pan, J.}, \bibinfo{author}{Alwala, K.V.}, \bibinfo{author}{Carion, N.}, \bibinfo{author}{Wu, C.Y.}, \bibinfo{author}{Girshick, R.}, \bibinfo{author}{Dollár, P.}, \bibinfo{author}{Feichtenhofer, C.}, \bibinfo{year}{2024}.
\newblock \bibinfo{title}{Sam 2: Segment anything in images and videos}.
\newblock \URLprefix \url{https://arxiv.org/abs/2408.00714}, \href{http://arxiv.org/abs/2408.00714}{\tt arXiv:2408.00714}.
%Type = Article
\bibitem[{Teobaldelli et~al.(2019)Teobaldelli, Rouphael, Fascella, Cristofori, Rivera and Basile}]{teobaldelli_developing_2019}
\bibinfo{author}{Teobaldelli, M.}, \bibinfo{author}{Rouphael, Y.}, \bibinfo{author}{Fascella, G.}, \bibinfo{author}{Cristofori, V.}, \bibinfo{author}{Rivera, C.M.}, \bibinfo{author}{Basile, B.}, \bibinfo{year}{2019}.
\newblock \bibinfo{title}{Developing an {Accurate} and {Fast} {Non}-{Destructive} {Single} {Leaf} {Area} {Model} for {Loquat} ({Eriobotrya} japonica {Lindl}) {Cultivars}}.
\newblock \bibinfo{journal}{Plants} \bibinfo{volume}{8}, \bibinfo{pages}{230}.
\newblock \URLprefix \url{https://www.mdpi.com/2223-7747/8/7/230}, \DOIprefix\doi{10.3390/plants8070230}. \bibinfo{note}{number: 7 Publisher: Multidisciplinary Digital Publishing Institute}.
%Type = Article
\bibitem[{Tian et~al.(2022)Tian, Ma, Yang and Duan}]{tian2022application}
\bibinfo{author}{Tian, Z.}, \bibinfo{author}{Ma, W.}, \bibinfo{author}{Yang, Q.}, \bibinfo{author}{Duan, F.}, \bibinfo{year}{2022}.
\newblock \bibinfo{title}{Application status and challenges of machine vision in plant factory—a review}.
\newblock \bibinfo{journal}{Information Processing in Agriculture} \bibinfo{volume}{9}, \bibinfo{pages}{195--211}.
%Type = Article
\bibitem[{Weraduwage et~al.(2015)Weraduwage, Chen, Anozie, Morales, Weise and Sharkey}]{weraduwage_relationship_2015}
\bibinfo{author}{Weraduwage, S.M.}, \bibinfo{author}{Chen, J.}, \bibinfo{author}{Anozie, F.C.}, \bibinfo{author}{Morales, A.}, \bibinfo{author}{Weise, S.E.}, \bibinfo{author}{Sharkey, T.D.}, \bibinfo{year}{2015}.
\newblock \bibinfo{title}{The relationship between leaf area growth and biomass accumulation in {Arabidopsis} thaliana}.
\newblock \bibinfo{journal}{Frontiers in Plant Science} \bibinfo{volume}{6}.
\newblock \URLprefix \url{https://www.frontiersin.org/journals/plant-science/articles/10.3389/fpls.2015.00167/full}, \DOIprefix\doi{10.3389/fpls.2015.00167}. \bibinfo{note}{publisher: Frontiers}.
%Type = Article
\bibitem[{Widuri et~al.(2017)Widuri, Lakitan, Hasmeda, Sodikin, Wijaya, Meihana, Kartika and Siaga}]{widuri_relative_2017}
\bibinfo{author}{Widuri, L.I.}, \bibinfo{author}{Lakitan, B.}, \bibinfo{author}{Hasmeda, M.}, \bibinfo{author}{Sodikin, E.}, \bibinfo{author}{Wijaya, A.}, \bibinfo{author}{Meihana, M.}, \bibinfo{author}{Kartika, K.}, \bibinfo{author}{Siaga, E.}, \bibinfo{year}{2017}.
\newblock \bibinfo{title}{Relative leaf expansion rate and other leaf-related indicators for detection of drought stress in chili pepper ('{Capsicum} annuum' {L}.)}.
\newblock \bibinfo{journal}{Australian Journal of Crop Science} \bibinfo{volume}{11}, \bibinfo{pages}{1617--1625}.
\newblock \URLprefix \url{https://search.informit.org/doi/abs/10.3316/INFORMIT.402390405327550}, \DOIprefix\doi{10.3316/informit.402390405327550}. \bibinfo{note}{publisher: Southern Cross Journals}.
%Type = Article
\bibitem[{Yau et~al.(2021)Yau, Ng and Lee}]{yau_portable_2021}
\bibinfo{author}{Yau, W.K.}, \bibinfo{author}{Ng, O.E.}, \bibinfo{author}{Lee, S.W.}, \bibinfo{year}{2021}.
\newblock \bibinfo{title}{Portable device for contactless, non-destructive and in situ outdoor individual leaf area measurement}.
\newblock \bibinfo{journal}{Computers and Electronics in Agriculture} \bibinfo{volume}{187}, \bibinfo{pages}{106278}.
\newblock \URLprefix \url{https://www.sciencedirect.com/science/article/pii/S0168169921002957}, \DOIprefix\doi{10.1016/j.compag.2021.106278}.
%Type = Article
\bibitem[{Zhou et~al.(2018)Zhou, Park and Koltun}]{Zhou2018open3d}
\bibinfo{author}{Zhou, Q.Y.}, \bibinfo{author}{Park, J.}, \bibinfo{author}{Koltun, V.}, \bibinfo{year}{2018}.
\newblock \bibinfo{title}{{Open3D}: {A} modern library for {3D} data processing}.
\newblock \bibinfo{journal}{arXiv:1801.09847} .

\end{thebibliography}

\bio{}
\endbio

% % Here goes the biography details.

\end{document}